\pdfoutput=1

\documentclass[11pt]{article}

\usepackage{acl}

\usepackage{times}
\usepackage{latexsym}
\usepackage{amsmath}
\usepackage{amssymb}

\usepackage{graphicx}

\usepackage[T1]{fontenc}

\usepackage[utf8]{inputenc}

\usepackage{microtype}

\usepackage{inconsolata}
\usepackage{kotex}
\usepackage{bm}
\usepackage{booktabs}
\usepackage[export]{adjustbox}
\usepackage{hhline}
\usepackage{multirow}
\usepackage{comment}

\usepackage{subcaption}
\usepackage{tabularx}

\usepackage[capitalize]{cleveref}
\crefname{section}{Sec.}{Secs.}
\Crefname{section}{Section}{Sections}
\Crefname{table}{Table}{Tables}
\crefname{table}{Tab.}{Tabs.}

\definecolor{azure}{rgb}{0.0, 0.5, 1.0}
\definecolor{awesome}{rgb}{1.0, 0.13, 0.32}
\definecolor{forestgreen}{rgb}{0.13, 0.55, 0.13}

\title{See It All: Contextualized Late Aggregation for 3D Dense Captioning}

\newcommand*\samethanks[1][\value{footnote}]{\footnotemark[#1]}
\author{
Minjung Kim\textsuperscript{\rm 1,\thanks{\hspace{0.2cm}Work done during internship at LG AI Research}}\hspace{0.4cm}
Hyung Suk Lim\textsuperscript{\rm 1}$^ ,$\textsuperscript{\rm 3}\hspace{0.4cm}
Seung Hwan Kim\textsuperscript{\rm 2}\hspace{0.4cm}
Soonyoung Lee\textsuperscript{\rm 2}\hspace{0.4cm} \vspace{0.2cm}\\
{\bf Bumsoo Kim\textsuperscript{\rm 2}$^ ,$\thanks{\hspace{0.2cm}Corresponding authors}}\hspace{0.4cm}
{\bf Gunhee Kim\textsuperscript{\rm 1}$^ ,$\samethanks}\vspace{0.2cm}
\\\textsuperscript{\rm 1}Seoul National University \hspace{0.4cm} \textsuperscript{\rm 2}LG AI Research \hspace{0.4cm} 
\textsuperscript{\rm 3}Diquest \\
{\tt\small \textsuperscript{\rm 1}minjung.kim@vision.snu.ac.kr, \textsuperscript{\rm 3}hslim@diquest.com, \textsuperscript{\rm 2}bumsoo.kim@lgresearch.ai, \textsuperscript{\rm 1}gunhee@snu.ac.kr}
}

\begin{document}
\maketitle
\begin{abstract}

3D dense captioning is a task to localize objects in a 3D scene and generate descriptive sentences for each object. Recent approaches in 3D dense captioning have adopted transformer encoder-decoder frameworks from object detection to build an end-to-end pipeline without hand-crafted components. However, these approaches struggle with contradicting objectives where a single query attention has to simultaneously view both the tightly localized object regions and contextual environment. To overcome this challenge, we introduce SIA (See-It-All), a transformer pipeline that engages in 3D dense captioning with a novel paradigm called late aggregation. SIA simultaneously decodes two sets of queries—context query and instance query. The instance query focuses on localization and object attribute descriptions, while the context query versatilely captures the region-of-interest of relationships between multiple objects or with the global scene, then aggregated afterwards (i.e., late aggregation) via simple distance-based measures. To further enhance the quality of contextualized caption generation, we design a novel aggregator to generate a fully informed caption based on the surrounding context, the global environment, and object instances. Extensive experiments on two of the most widely-used 3D dense captioning datasets demonstrate that our proposed method achieves a significant improvement over prior methods.

\end{abstract}

\section{Introduction}
\label{sec:intro}
\begin{figure*}[t]
\begin{center}
\centerline{\includegraphics[width=0.88\textwidth]{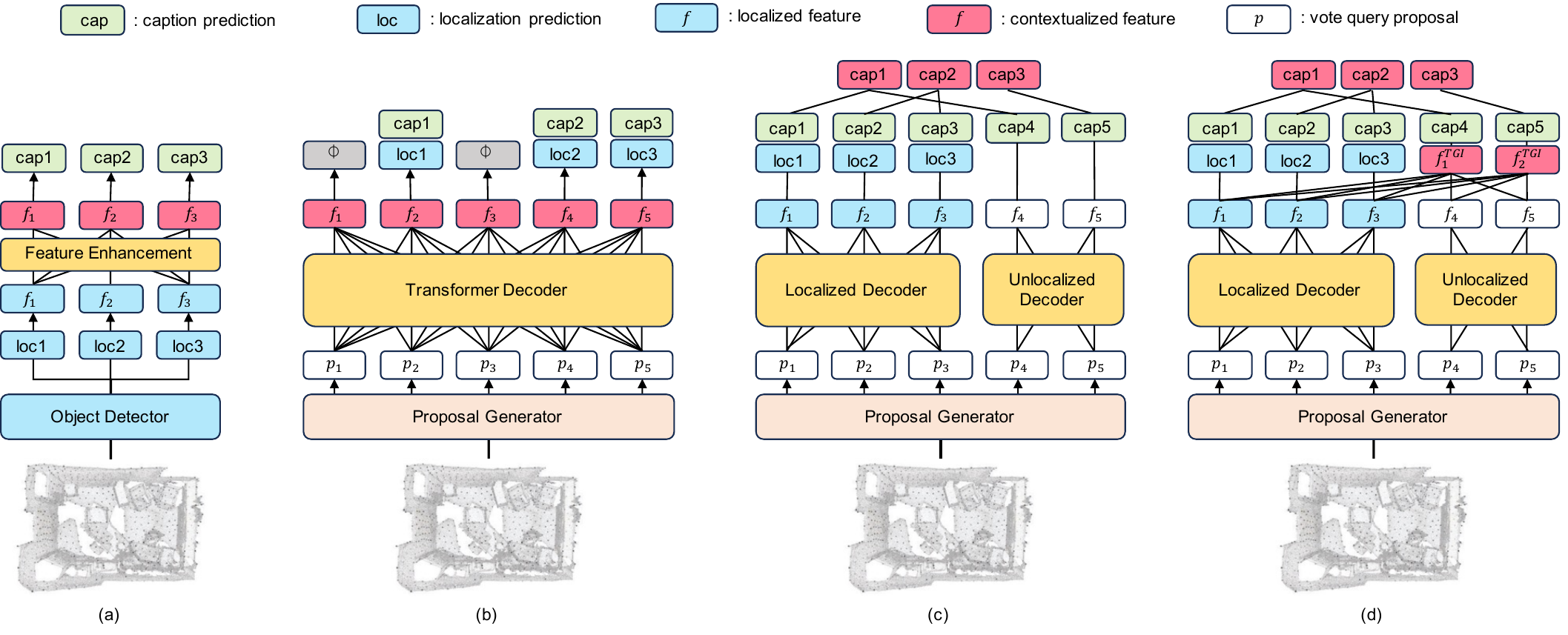}}
\caption{
Schematic diagrams illustrating paradigms of 3D dense captioning: (a) features are extracted from object detectors, and their relations are further aggregated to enhance features~\citet{cai20223djcg} (b) proposals are generated by voting, then the local-context features are aggregated with transformer attention~\citet{chen2023vote2capdetr} (c) our proposed SIA separately encodes features with local boundaries and context features without such boundaries, and aggregates the generated caption that involves identical objects afterward (i.e., \textit{late aggregation}) (d) SIA with further enhanced contextual features generated from our novel TGI-Aggregator ($f^{TGI}$) that aggregates local-context-global features for a more contextualized caption generation.
}
\label{fig:teaser}
\end{center}
\end{figure*}


3D dense captioning has been defined in former works~\cite{chen2021scan2cap, chen2022d3net, yuan2022xtrans2cap, wang2022spacap3d, jiao2022more, cai20223djcg, chen2023vote2capdetr} as the task of localizing all the objects in a 3D scene (i.e., object detection) and generating descriptive sentences for each object (i.e., object caption generation).
Early works incorporated a two-stage ``detect-then-describe" pipeline, where we first detect all the object proposals then generate the captions for each object~\cite{chen2021scan2cap, jiao2022more, wang2022spacap3d, zhong2022contextual, cai20223djcg, chen2022d3net, yuan2022xtrans2cap}.
However, the sequential design, lacking sufficient integration of contextual information in these endeavors, has been limited in performance and efficiency.

Vote2Cap-DETR~\cite{chen2023vote2capdetr} has emulated the transformer encoder-decoder pipeline from object detection~\cite{carion2020detr} to alleviate these issues and fashioned an end-to-end pipeline for 3D dense captioning.
Powered by transformer attentions, this method contextualizes individual objects (i.e., self-attention with other proposals throughout the global scene) to generate dense captions.
Nevertheless, compared to the notable advancements that object detection has experienced, the direct application of this architecture has failed to fully leverage the contextual information required for 3D dense captioning.

Dense captioning has to perform precise object localization while generating captions that either independently describe an object's attributes (e.g., a \textit{wooden} chair) or describe the object within its contextual environment (e.g., a chair \textit{in front of} the TV).
This presents a challenging scenario where the feature representation for a single query must encompass both accurate local features for localization or attribute-based caption generation, alongside incorporating contextual features that dynamically span neighboring regions or the broader global scene.
Focusing attention on local features can enhance localization and detailed attribute description but reduce sensitivity to the surrounding context. 
Conversely, spreading attention to include the context can improve understanding of the environmental description but at the cost of localization accuracy.

In this paper, we propose a pipeline engaging a novel \textit{late aggregation} paradigm called \textbf{SIA (i.e., See It All)}.
Rather than assigning a query \textit{per object} and training them to dynamically incorporate local features, contextual information, and the global scene, SIA allocates a query \textit{per caption}.
To elaborate, SIA identifies a distinct region to focus on when generating each caption and then consolidates the outcomes concerning identical objects.
Figure~\ref{fig:teaser} contrasts our proposed late aggregation approach with existing dense captioning paradigms.
While previous works have either (a) extracted the features from localized object areas~\citet{cai20223djcg} or (b) generated captions from features that have to perform both localization and proper caption prediction~\citet{chen2023vote2capdetr}, SIA (c) focuses on each unique ROIs for each caption then aggregates the captions that include identical objects afterwards.
This architecture (i.e., \textit{late aggregation}) enables SIA to produce attribute captions and localization with features concentrated exclusively on specific local areas, while captions necessitating a broad range of contextual information can be crafted using features gathered without the constraints of localization boundaries.

To further refine the features for contextual captions, we design a unique aggregator that generates captions based on the con\textbf{T}extual surroundings, \textbf{G}lobal descriptor, and \textbf{I}nstance features (i.e., TGI-Aggregator, see \Cref{fig:teaser}-(d)).
Our TGI-Aggregator generates contextual caption based on the fully informed feature that can dynamically capture the area of interest within the scene without the constraints of localization objectives. 
Extensive experiments on two widely used benchmarks in 3D dense captioning (i.e., ScanRefer~\cite{chen2020scanrefer} and Nr3D~\cite{achlioptas2020referit3d}) show that our proposed SIA surpasses prior approaches by a large margin. 
The contribution of our paper can be summarized as:
\begin{itemize}
    \item We propose a new paradigm for 3D dense captioning (i.e., \textit{late aggregation}). While previous works aggregate the instance and context features first and then generate captions, SIA generates local and contextual captions separately and then aggregates the captions involving identical objects. 
    \item To further improve the quality of features used for contextualized caption generation, we propose a novel aggregator named TGI-Aggregator.
    \item Our SIA achieves state-of-the-art performances across multiple evaluation metrics on the ScanRefer and Nr3D datasets.
\end{itemize}

\section{Preliminary}
\label{sec:preliminary}

In this preliminary, we start with a basic transformer-based end-to-end 3D dense captioning pipeline~\cite{chen2023vote2capdetr}.
The caption head is attached to the top of the existing 3D object detection pipeline~\cite{votenet, 3detr} with vote queries that establish captions for each pinpointed object throughout the scene in an object-by-object manner.
Afterward, we discuss why this object-centric application of transformer attention is unsuitable for 3D dense captioning.

\begin{figure*}[t]
\begin{center}
\centerline{\includegraphics[width=.95\textwidth]{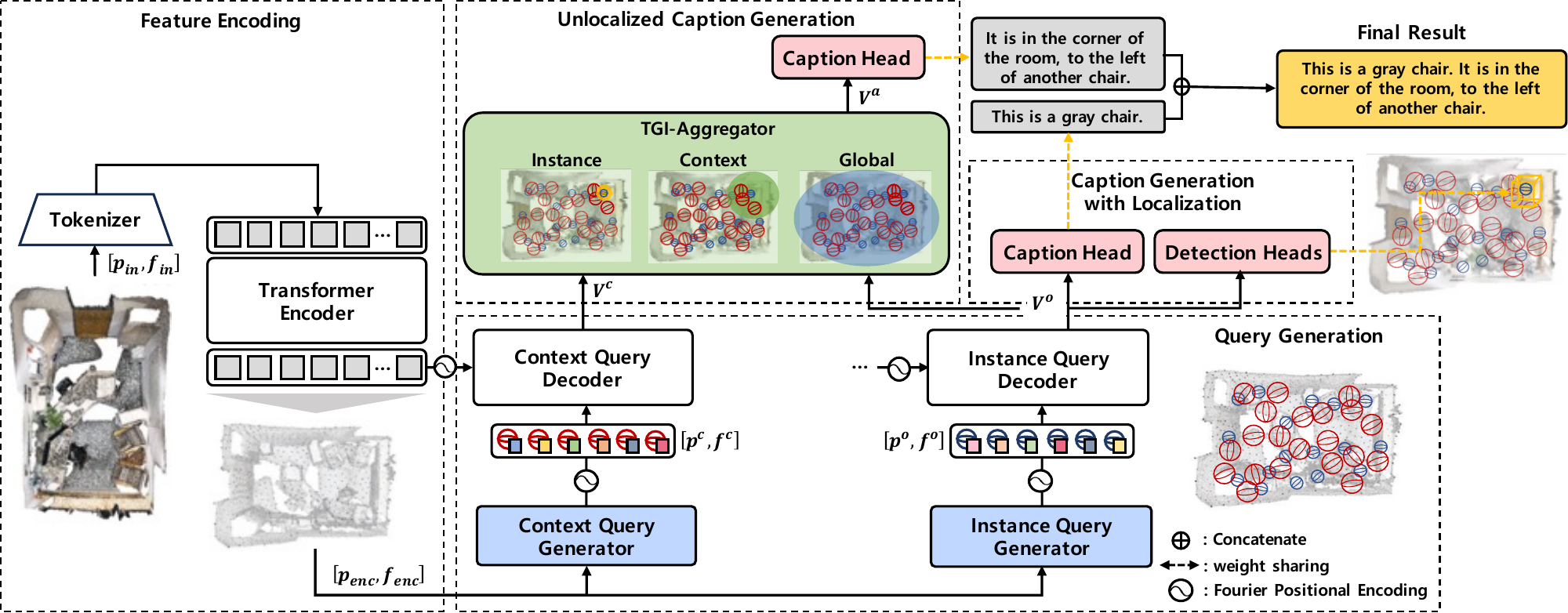}}
\caption{
Overall architecture of SIA for 3D dense captioning. The caption query set is each designated to Instance Query Decoder and Context Query Decoder. In the Instance Query Decoder, the caption based on the tight localized area are generated along with object detection. In the Context Query Decoder, captions that require views transcending single object localization such as captions containing relation between multiple objects or relation between the scene are generated. The feature for this Unlocalized Caption Generation is further enhanced with our novel TGI-Aggregator, that contextualizes the feature from con\textbf{T}ext regions, the \textbf{G}lobal scene, and \textbf{I}nstances.
}
\label{fig:overall}
\end{center}
\end{figure*}
\subsection{End-to-End 3D Object Detection}

3D object detection aims to identify and localize objects in 3D scenes.
VoteNet~\cite{votenet} incorporates an encoder-decoder architecture where the bounding boxes are predicted by aggregating the \textit{votes} for the center coordinates.
3DETR~\cite{3detr} generates \textit{object queries} by uniformly sampling seed points from a 3D scene.
Vote2Cap-DETR~\cite{chen2023vote2capdetr} uses \textit{vote queries} that connect the object queries in 3DETR to VoteNet, resulting in better localization and improved training efficiencies.
\subsection{Extension to 3D Dense Captioning}

The goal of 3D dense captioning is to localize objects in a 3D scene and generate informative natural language descriptions per object.
An intuitive extension from object detection to 3D dense captioning is simply applying a captioning head for each object proposals~\cite{chen2023vote2capdetr}.
Given an input indoor 3D scene as a point cloud $PC=[p_{\text{in}};f_{\text{in}}]\in\mathbb{R}^{N\times(3+F)}$, where $p_{\text{in}}\in\mathbb{R}^{N\times3}$ is the absolute locations for each point and $f_{\text{in}}\in\mathbb{R}^{N\times F}$ is additional input features for each point~\cite{chen2020scanrefer,chen2021scan2cap}, the objective of 3D dense captioning is to generate a set of box-caption pairs $(\hat{B},\hat{C})=\{(\hat{b_1},\hat{c_1}),...,(\hat{b_K},\hat{c_K})\}$, representing an estimation of $K$ distinctive objects in this 3D scene.
Captions are generated in \textit{parallel} with bounding box prediction using a caption head.
Since the aforementioned vote queries (i.e., $p_{\text{vote}}$) fail to provide adequate attributes and spatial relations for informative caption generation, the contextual information is leveraged through a separate lightweight transformer~\cite{chen2023vote2capdetr}.
\subsection{Retrospect on Object-Centric Captioning}

Current 3D dense captioning benchmarks require the model to generate multiple captions for each detected object.
Therefore, it seems natural to approach this task in an \textit{object-centric} manner~\cite{wang2022spacap3d, jiao2022more, achlioptas2020referit3d}, where we generate captions per each object proposal.
However, unlike object detection, dense captioning requires an extensive understanding of the scene, including the attributes of each object and the relative information between objects and the global scene.
Therefore, designating the queries per object requires a single query attention to versatilely encompass the individual object and its surrounding elements, failing to concentrate on the local element it should describe effectively.
We propose a novel \textit{late aggregation} approach for 3D dense captioning to address this issue and incorporate contextual scene information.
\section{Method}
\label{sec:method}

In this section, we introduce a transformer encoder-decoder pipeline that engages our novel \textit{late aggregation} paradigm for 3D dense captioning.
In previous methods, the transformer attention aggregates contextual information \textit{per-object}, where a single feature is used to perform localization, generate localized attributes and simultaneously capture the surrounding context area. 
SIA is designed to capture the unique region of interest for each caption.
Local attribute descriptions are generated with localized features. In contrast, contextualized captions that include relationships with other objects or the entire scene are generated with a separately decoded feature irrelevant to localization objectives.
Then, captions involving identical objects are aggregated via distance (i.e., \textit{late aggregation}) to consist of the final caption.
The overall pipeline is illustrated in \Cref{fig:overall}.

\subsection{Encoder}

Given the input point cloud $PC=[p_{\text{in}};f_{\text{in}}]\in\mathbb{R}^{N\times(3+F)}$, the input point cloud is first tokenized by a set-abstraction layer of PointNet++~\cite{qi2017pointnet++}.
The tokenized output is inputted into a masked transformer encoder with the set-abstraction layer, followed by two additional encoder layers.
The final encoded scene tokens are denoted as $p_{enc}\in\mathbb{R}^{1,024\times 3}$ and $f_{enc}\in\mathbb{R}^{1,024\times 256}$. 
\subsection{Context Query and Instance Query}

To disentangle the captions that are bound to a single object and captions that include relative information with other objects or the global scene, we designate two separate \textit{instance query} and \textit{context query} to each capture a unique region per caption within the 3D scene.
While the context query captures the local-global regions capable of captioning, the \textit{instance query} generates standard object localization and attribute-related caption prediction for each object.
The two queries are decoded in parallel and later aggregated to consist of the final caption.

\paragraph{Context Query Generator.}
Given the encoded scene tokens $(p_{\text{enc}}, f_{\text{enc}})$, we sample $512$ context points $p^c_{\text{seed}}$ with farthest point sampling (FPS) on $p_{\text{enc}}$. 
Then, the context query $(p^c,f^c)$ is represented as:
\begin{equation}
    (p^c,f^c)=\text{SA}_c(p_{\text{enc}},f_{\text{enc}}),
\end{equation}
where $\text{SA}_c$ denotes the set-abstraction layer~\cite{qi2017pointnet++} with a radius of $1.2$ and samples $64$ points for $p^c$.

\paragraph{Instance Query Generator.}
The instance query is decoded to perform standard 3D object detection and generate captions for the individual attributes of each object.
Likewise, the instance query $(p^o,f^o)$ is written as:
\begin{equation}
    [\Delta p_{\text{vote}}; \Delta f_\text{vote}]=\text{FFN}_{o}(f_{\text{enc}}),
\end{equation}
\begin{equation}
    (p^o,f^o)=\text{SA}_o\big(p_{\text{enc}}+\Delta p_{\text{vote}}, f_\text{enc}+\Delta f_\text{vote}\big),
\end{equation}
where $[\Delta p_{\text{vote}}; \Delta f_\text{vote}]\in\mathbb{R}^{1,024\times(3+256)}$ is an offset that learns to shift the encoded points to object's centers spatially by a feed-forward network $\text{FFN}_{o}$, following \cite{chen2023vote2capdetr}.
$\text{SA}_o$ denotes the set-abstraction layer with a radius of $0.3$ and samples $16$ points for $p^o$.
All hyper-parameters are set experimentally.
\subsection{Decoding}

Given the context and instance queries, we build a parallel decoding pipeline where the Context Decoder describes contextual information between objects, and the Instance Decoder performs the localization and attribute description.
We then feed the decoded context query $V^c$ and instance query $V^o$ to our TGI-Aggregator.

\paragraph{TGI-Aggregator.}
\begin{figure}[t]
\begin{center}
\centerline{\includegraphics[width=.95\columnwidth]{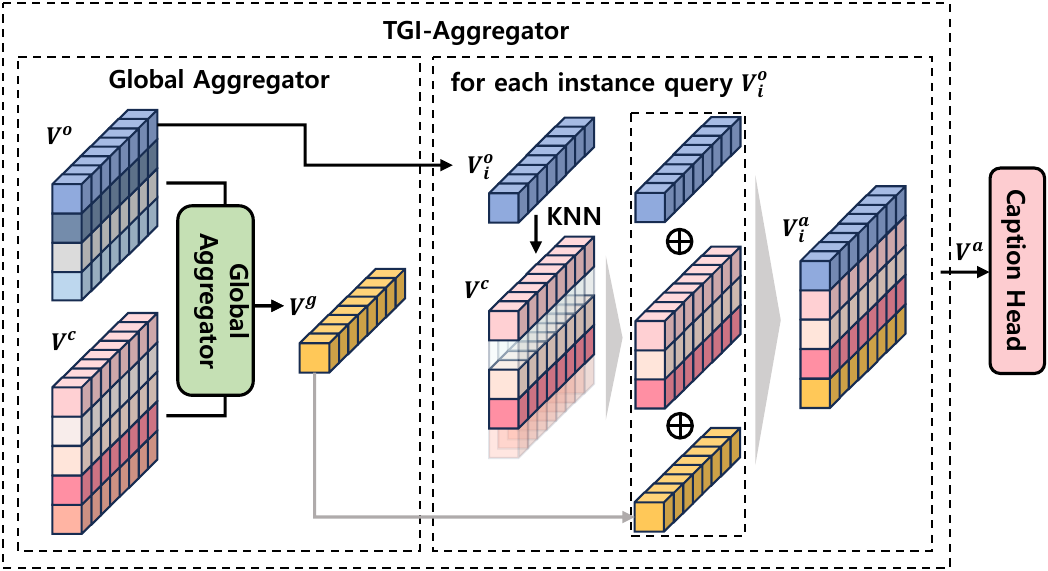}}
\caption{
Conceptual illustration of our TGI-Aggregator. The Global Aggregator $G(\cdot)$ aggregates the decoded context query $V^o$ and instance query $V^c$ to construct a global descriptor $V^g$. Then, the instance feature $V_i^o$, the nearest neighbor feature in $V^c$, and the global descriptor $V^g$ are concatenated to construct $V^a$.
}
\vspace{-15pt}
\label{fig:tgi}
\end{center}
\end{figure}
\Cref{fig:tgi} shows a conceptual illustration of our TGI-Aggregator.
To encompass the understanding of the entire scene for each caption, we generate a global feature using all decoded context queries $V^c$ and instance queries $V^o$; related experiments can be found in \Cref{sec:ablation_study}.
We deploy a clustering-based aggregator~\cite{netvlad} $G(\cdot)$.
As a result, we obtain a global descriptor $V^g\in\mathbb{R}^{256}$ by $V^g=G(V^c,V^o)$.
Then, we concatenate this global descriptor $V^g$ to each decoded instance query $V^o_i$ and $K$ nearest features within $V^c$ in terms of spatial proximity to $V^o_i$, resulting in an aggregated feature $V^{a}$ that contains a comprehensive information of con\textbf{T}ext, \textbf{G}lobal, and \textbf{I}nstance.
We set $K$ to $16$.

\paragraph{Contextual Caption Generation.}
For caption generation, we adopt a transformer decoder-based caption head based on GPT-2~\cite{gpt2}, following Vote2Cap-DETR~\cite{chen2023vote2capdetr} and SpaCap3D~\cite{wang2022spacap3d}.
The output $V^{a}$ of our TGI-Aggregator contains features about contextual surroundings, global context, and instance information.
Based on this feature, SIA can generate descriptions for relationships between multiple objects (e.g., the chair is next to a bookshelf) and global relationships (e.g., the table is in the middle of the room).

\paragraph{Localization \& Attribute Caption Generation.}
To predict localization and attribute descriptions for the participating instances, we feed the decoded instance query $V^o$ and parallelly feed it into the detection head and (shared) caption head.
For localization, we follow 3DETR~\cite{3detr}, reformulating the box corner estimation as offset estimation from a query point to an object’s center and box size regression.
All subtasks are implemented by FFNs.
The object localization head is shared throughout the decoder layers. 
\subsection{Training SIA}
We construct the final caption to be object-centric to compare with previous object-centric methods using benchmark datasets.
The final caption for the $i$-th object is obtained by simply concatenating the captions generated from  $V^o_i$ and $V^a_i$.
Our SIA is trained and evaluated by locating all objects within a scene and comparing the final caption centered on each object with the ground-truth.

\paragraph{Instance Query Loss.}
To train the Instance Query Generator to find an object's center by shifting points $p_{enc}$, we adopt the vote loss from VoteNet~\cite{votenet}.
Given the generated instance query $(p^o,f^o)$ and the encoded scene tokens $(p_{enc},f_{enc})$, the vote loss $\mathcal{L}^o$ is written as:
\begin{equation}
    \mathcal{L}^o=\frac{1}{M}\sum_{i=1}^M\sum_{j=1}^{N_{gt}}{\Vert p^o_i-{\text{cnt}}_j\Vert_1\cdot \mathbb{I}(p^i_{enc})},
\end{equation}
where $\mathbb{I}(x)$ is an indicator function that equals 1 when $x\in I_j$ and 0 otherwise, $N_{gt}$ is the number of instances in a 3D scene, $M$ is the number of $p^o$, and ${\text{cnt}}_j$ is the
center of $j$-th instance $I_j$.

\paragraph{Detection Loss.}
We use Hungarian matching~\cite{kuhn1955hungarian} to assign each proposal with the ground-truth, following DETR~\cite{carion2020detr}.
The detection loss $\mathcal{L}_{\text{det}}$ is written as:
\begin{equation}
\mathcal{L}_{\text{det}}=\alpha_1\mathcal{L}_{\text{giou}}+\alpha_2\mathcal{L}_{\text{cls}}+\alpha_3\mathcal{L}_{\text{cnt}}+\alpha_4\mathcal{L}_{\text{size}},
\end{equation}
where $\alpha_1=10, \alpha_2=1, \alpha_3=5, \alpha_4=1$ are set heuristically.
The detection loss is applied across all decoder layers for better convergence.

\paragraph{Caption Loss.}
Following the standard protocol for image captioning, we first train caption heads with standard cross-entropy loss for Maximum Likelihood Estimation (MLE).
In the MLE training, the model learns to predict the $(t+1)$-th word $c^{t+1}_n$ based on the first $t$ words $c^{[1:t]}_n$ and the visual features $\mathcal{V}$.
The loss function is established for the final caption with length $T$ is defined as follows:
\begin{equation}
    \mathcal{L}_{c_n}=\sum_{t=1}^T\mathcal{L}_{c_n}(t)=-\sum_{t=1}^T\log{\hat{P}\bigg(c_{n}^{t+1}\vert\mathcal{V},c_{n}^{[1:t]}\bigg)},
\end{equation}
Once the caption head is trained with word-level supervision, it is refined using Self-Critical Sequence Training (SCST)~\cite{scst}.
In this phase, the model produces multiple captions $\hat{c}_{1,...,k}$ with a beam size of $k$ and an additional $\hat{g}$ using greedy search as a baseline. 
The loss function for SCST is formulated as follows:
\begin{equation}
    \mathcal{L}_{c_n}=-\sum_{i=1}^k(R(\hat{c_i})-R(\hat{g}))\cdot\frac{1}{\vert\hat{c_i}\vert}\log{\hat{P}(\hat{c}_i\vert\mathcal{V})}.
\end{equation}
The reward function $R(\cdot)$ is based on the CIDEr~\cite{vedantam2015cider} metric for caption evaluation, and the logarithmic probability of the caption $\hat{c}_i$ is normalized by caption length $|\hat{c}_i|$, promoting equal importance to captions of varying lengths by the model.

\paragraph{Final Loss for SIA.}
Given the instance query Loss $\mathcal{L}^o$, the detection loss for the $i$-th decoder layer as $\mathcal{L}^i_{\text{det}}$, and the average of the caption loss $\mathcal{L}_{c_n}$ within a batch denoted as $\mathcal{L}_{\text{cap}}$, the final loss $\mathcal{L}$ for SIA is written as:
\begin{equation}
\mathcal{L}=\beta_1 \mathcal{L}^o+\beta_2 \sum_{i=1}^{n_{\text {dec-layer }}} \mathcal{L}^i_{\text {det}}+\beta_3 \mathcal{L}_{\text {cap }},
\end{equation}
where $\beta_{1}=10$, $\beta_{2}=1$, and $\beta_{3}=10$.
\begin{table*}[t]
   \centering
   \resizebox{\textwidth}{!}{%
     \begin{tabular}{c@{\hspace{-2pt}}c|cccc@{\hspace{30pt}}cccc@{\hspace{30pt}}cccc@{\hspace{30pt}}cccc}
     \toprule
     \multirow{1}{*}{} & \multicolumn{1}{c}{} & \multicolumn{8}{c}{w/o additional 2D data} & \multicolumn{8}{c}{w/ additional 2D data} \\
     \multirow{1}{*}{Model} & \multicolumn{1}{c}{Training} & \multicolumn{4}{c}{IoU=$0.25$} & \multicolumn{4}{c}{IoU=$0.50$} & \multicolumn{4}{c}{IoU=$0.25$} & \multicolumn{4}{c}{IoU=$0.50$} \\
     \cmidrule(lr){3-6} \cmidrule(lr){7-10} \cmidrule(lr){11-14} \cmidrule(lr){15-18}
     \multirow{1}{*}{} & \multicolumn{1}{c}{} 
     & \multicolumn{1}{c}{C$\uparrow$} & \multicolumn{1}{c}{B-4$\uparrow$} & \multicolumn{1}{c}{M$\uparrow$} & \multicolumn{1}{c}{R$\uparrow$} 
     & \multicolumn{1}{c}{C$\uparrow$} & \multicolumn{1}{c}{B-4$\uparrow$} & \multicolumn{1}{c}{M$\uparrow$} & \multicolumn{1}{c}{R$\uparrow$} 
     & \multicolumn{1}{c}{C$\uparrow$} & \multicolumn{1}{c}{B-4$\uparrow$} & \multicolumn{1}{c}{M$\uparrow$} & \multicolumn{1}{c}{R$\uparrow$} 
     & \multicolumn{1}{c}{C$\uparrow$} & \multicolumn{1}{c}{B-4$\uparrow$} & \multicolumn{1}{c}{M$\uparrow$} & \multicolumn{1}{c}{R$\uparrow$} \\
     \midrule
     \midrule
     
     \multirow{1}{*}{Scan2Cap} & \multicolumn{1}{c}{} 
     & \multicolumn{1}{c}{53.73} & \multicolumn{1}{c}{34.25} & \multicolumn{1}{c}{26.14} & \multicolumn{1}{c}{54.95} 
     & \multicolumn{1}{c}{35.20} & \multicolumn{1}{c}{22.36} & \multicolumn{1}{c}{21.44} & \multicolumn{1}{c}{43.57} 
     & \multicolumn{1}{c}{56.82} & \multicolumn{1}{c}{34.18} & \multicolumn{1}{c}{26.29} & \multicolumn{1}{c}{55.27}
     & \multicolumn{1}{c}{39.08} & \multicolumn{1}{c}{23.32} & \multicolumn{1}{c}{21.97} & \multicolumn{1}{c}{44.78} \\

     \multirow{1}{*}{D3Net} & \multicolumn{1}{c}{} 
     & \multicolumn{1}{c}{-} & \multicolumn{1}{c}{-} & \multicolumn{1}{c}{-} & \multicolumn{1}{c}{-} 
     & \multicolumn{1}{c}{-} & \multicolumn{1}{c}{-} & \multicolumn{1}{c}{-} & \multicolumn{1}{c}{-} 
     & \multicolumn{1}{c}{-} & \multicolumn{1}{c}{-} & \multicolumn{1}{c}{-} & \multicolumn{1}{c}{-}
     & \multicolumn{1}{c}{46.07} & \multicolumn{1}{c}{30.29} & \multicolumn{1}{c}{24.35} & \multicolumn{1}{c}{51.67} \\

     \multirow{1}{*}{SpaCap3d} & \multicolumn{1}{c}{} 
     & \multicolumn{1}{c}{58.06} & \multicolumn{1}{c}{35.30} & \multicolumn{1}{c}{26.16} & \multicolumn{1}{c}{55.03} 
     & \multicolumn{1}{c}{42.76} & \multicolumn{1}{c}{25.38} & \multicolumn{1}{c}{22.84} & \multicolumn{1}{c}{45.66} 
     & \multicolumn{1}{c}{63.30} & \multicolumn{1}{c}{36.46} & \multicolumn{1}{c}{26.71} & \multicolumn{1}{c}{55.71}
     & \multicolumn{1}{c}{44.02} & \multicolumn{1}{c}{25.26} & \multicolumn{1}{c}{22.33} & \multicolumn{1}{c}{45.36} \\

     \multirow{1}{*}{MORE} & \multicolumn{1}{c}{} 
     & \multicolumn{1}{c}{58.89} & \multicolumn{1}{c}{35.41} & \multicolumn{1}{c}{26.36} & \multicolumn{1}{c}{55.41} 
     & \multicolumn{1}{c}{38.98} & \multicolumn{1}{c}{23.01} & \multicolumn{1}{c}{21.65} & \multicolumn{1}{c}{44.33} 
     & \multicolumn{1}{c}{62.91} & \multicolumn{1}{c}{36.25} & \multicolumn{1}{c}{26.75} & \multicolumn{1}{c}{56.33}
     & \multicolumn{1}{c}{40.94} & \multicolumn{1}{c}{22.93} & \multicolumn{1}{c}{21.66} & \multicolumn{1}{c}{44.42} \\

     \multirow{1}{*}{3DJCG} & \multicolumn{1}{c}{} 
     & \multicolumn{1}{c}{60.86} & \multicolumn{1}{c}{39.67} & \multicolumn{1}{c}{27.45} & \multicolumn{1}{c}{59.02} 
     & \multicolumn{1}{c}{47.68} & \multicolumn{1}{c}{31.53} & \multicolumn{1}{c}{24.28} & \multicolumn{1}{c}{51.80} 
     & \multicolumn{1}{c}{64.70} & \multicolumn{1}{c}{40.17} & \multicolumn{1}{c}{27.66} & \multicolumn{1}{c}{59.23}
     & \multicolumn{1}{c}{49.48} & \multicolumn{1}{c}{31.03} & \multicolumn{1}{c}{24.22} & \multicolumn{1}{c}{50.80} \\

     \multirow{1}{*}{Contextual} & \multicolumn{1}{c}{MLE} 
     & \multicolumn{1}{c}{-} & \multicolumn{1}{c}{-} & \multicolumn{1}{c}{-} & \multicolumn{1}{c}{-}
     & \multicolumn{1}{c}{42.77} & \multicolumn{1}{c}{23.60} & \multicolumn{1}{c}{22.05} & \multicolumn{1}{c}{45.13}
     & \multicolumn{1}{c}{-} & \multicolumn{1}{c}{-} & \multicolumn{1}{c}{-} & \multicolumn{1}{c}{-}
     & \multicolumn{1}{c}{46.11} & \multicolumn{1}{c}{25.47} & \multicolumn{1}{c}{22.64} & \multicolumn{1}{c}{45.96} \\

     \multirow{1}{*}{REMAN} & \multicolumn{1}{c}{} 
     & \multicolumn{1}{c}{-} & \multicolumn{1}{c}{-} & \multicolumn{1}{c}{-} & \multicolumn{1}{c}{-}
     & \multicolumn{1}{c}{-} & \multicolumn{1}{c}{-} & \multicolumn{1}{c}{-} & \multicolumn{1}{c}{-} 
     & \multicolumn{1}{c}{62.01} & \multicolumn{1}{c}{36.37} & \multicolumn{1}{c}{27.76} & \multicolumn{1}{c}{56.25}
     & \multicolumn{1}{c}{45.00} & \multicolumn{1}{c}{26.31} & \multicolumn{1}{c}{22.67} & \multicolumn{1}{c}{46.96} \\
     
     \multirow{1}{*}{3D-VLP} & \multicolumn{1}{c}{} 
     & \multicolumn{1}{c}{64.09} & \multicolumn{1}{c}{39.84} & \multicolumn{1}{c}{27.65} & \multicolumn{1}{c}{58.78}
     & \multicolumn{1}{c}{50.02} & \multicolumn{1}{c}{31.87} & \multicolumn{1}{c}{24.53} & \multicolumn{1}{c}{51.17} 
     & \multicolumn{1}{c}{70.73} & \multicolumn{1}{c}{41.03} & \multicolumn{1}{c}{28.14} & \multicolumn{1}{c}{59.72}
     & \multicolumn{1}{c}{54.94} & \multicolumn{1}{c}{32.31} & \multicolumn{1}{c}{24.83} & \multicolumn{1}{c}{51.51} \\

     \multirow{1}{*}{Vote2Cap-DETR} & \multicolumn{1}{c}{} 
     & \multicolumn{1}{c}{71.45} & \multicolumn{1}{c}{39.34} & \multicolumn{1}{c}{28.25} & \multicolumn{1}{c}{59.33}
     & \multicolumn{1}{c}{61.81} & \multicolumn{1}{c}{34.46} & \multicolumn{1}{c}{26.22} & \multicolumn{1}{c}{54.40} 
     & \multicolumn{1}{c}{72.79} & \multicolumn{1}{c}{39.17} & \multicolumn{1}{c}{28.06} & \multicolumn{1}{c}{59.23}
     & \multicolumn{1}{c}{59.32} & \multicolumn{1}{c}{32.42} & \multicolumn{1}{c}{25.28} & \multicolumn{1}{c}{52.38} \\

     \multirow{1}{*}{Unit3D} & \multicolumn{1}{c}{} 
     & \multicolumn{1}{c}{-} & \multicolumn{1}{c}{-} & \multicolumn{1}{c}{-} & \multicolumn{1}{c}{-}
     & \multicolumn{1}{c}{-} & \multicolumn{1}{c}{-} & \multicolumn{1}{c}{-} & \multicolumn{1}{c}{-}
     & \multicolumn{1}{c}{-} & \multicolumn{1}{c}{-} & \multicolumn{1}{c}{-} & \multicolumn{1}{c}{-}
     & \multicolumn{1}{c}{46.69} & \multicolumn{1}{c}{27.22} & \multicolumn{1}{c}{21.91} & \multicolumn{1}{c}{45.98} \\

     \multirow{1}{*}{\textbf{Ours}} & \multicolumn{1}{c}{} 
     & \multicolumn{1}{c}{\textbf{78.68}} & \multicolumn{1}{c}{\textbf{43.25}} & \multicolumn{1}{c}{\textbf{29.21}} & \multicolumn{1}{c}{\textbf{63.06}} 
     & \multicolumn{1}{c}{\textbf{73.22}} & \multicolumn{1}{c}{\textbf{40.91}} & \multicolumn{1}{c}{\textbf{28.19}} & \multicolumn{1}{c}{\textbf{60.46}} 
     & \multicolumn{1}{c}{\textbf{78.05}} & \multicolumn{1}{c}{\textbf{42.16}} & \multicolumn{1}{c}{\textbf{28.74}} & \multicolumn{1}{c}{\textbf{61.70}}
     & \multicolumn{1}{c}{\textbf{69.86}} & \multicolumn{1}{c}{\textbf{37.89}} & \multicolumn{1}{c}{\textbf{27.04}} & \multicolumn{1}{c}{\textbf{57.33}} \\
     
     \midrule

     \multirow{1}{*}{Scan2Cap} & \multicolumn{1}{c}{} 
     & \multicolumn{1}{c}{-} & \multicolumn{1}{c}{-} & \multicolumn{1}{c}{-} & \multicolumn{1}{c}{-} 
     & \multicolumn{1}{c}{-} & \multicolumn{1}{c}{-} & \multicolumn{1}{c}{-} & \multicolumn{1}{c}{-} 
     & \multicolumn{1}{c}{-} & \multicolumn{1}{c}{-} & \multicolumn{1}{c}{-} & \multicolumn{1}{c}{-}
     & \multicolumn{1}{c}{48.38} & \multicolumn{1}{c}{26.09} & \multicolumn{1}{c}{22.15} & \multicolumn{1}{c}{44.74} \\

     \multirow{1}{*}{D3Net} & \multicolumn{1}{c}{} 
     & \multicolumn{1}{c}{-} & \multicolumn{1}{c}{-} & \multicolumn{1}{c}{-} & \multicolumn{1}{c}{-} 
     & \multicolumn{1}{c}{-} & \multicolumn{1}{c}{-} & \multicolumn{1}{c}{-} & \multicolumn{1}{c}{-} 
     & \multicolumn{1}{c}{-} & \multicolumn{1}{c}{-} & \multicolumn{1}{c}{-} & \multicolumn{1}{c}{-}
     & \multicolumn{1}{c}{62.64} & \multicolumn{1}{c}{35.68} & \multicolumn{1}{c}{25.72} & \multicolumn{1}{c}{53.90} \\

     \multirow{1}{*}{$\chi$-Tran2Cap} & \multicolumn{1}{c}{} 
     & \multicolumn{1}{c}{58.81} & \multicolumn{1}{c}{34.17} & \multicolumn{1}{c}{25.81} & \multicolumn{1}{c}{54.10} 
     & \multicolumn{1}{c}{41.52} & \multicolumn{1}{c}{23.83} & \multicolumn{1}{c}{21.90} & \multicolumn{1}{c}{44.97} 
     & \multicolumn{1}{c}{61.83} & \multicolumn{1}{c}{35.65} & \multicolumn{1}{c}{26.61} & \multicolumn{1}{c}{54.70}
     & \multicolumn{1}{c}{43.87} & \multicolumn{1}{c}{25.05} & \multicolumn{1}{c}{22.46} & \multicolumn{1}{c}{45.28} \\

     \multirow{1}{*}{Contextual} & \multicolumn{1}{c}{SCST} 
     & \multicolumn{1}{c}{-} & \multicolumn{1}{c}{-} & \multicolumn{1}{c}{-} & \multicolumn{1}{c}{-}
     & \multicolumn{1}{c}{50.29} & \multicolumn{1}{c}{25.64} & \multicolumn{1}{c}{22.57} & \multicolumn{1}{c}{44.71} 
     & \multicolumn{1}{c}{-} & \multicolumn{1}{c}{-} & \multicolumn{1}{c}{-} & \multicolumn{1}{c}{-}
     & \multicolumn{1}{c}{54.30} & \multicolumn{1}{c}{27.24} & \multicolumn{1}{c}{23.30} & \multicolumn{1}{c}{45.81} \\

     \multirow{1}{*}{Vote2Cap-DETR} & \multicolumn{1}{c}{} 
     & \multicolumn{1}{c}{84.15} & \multicolumn{1}{c}{42.51} & \multicolumn{1}{c}{28.47} & \multicolumn{1}{c}{59.26} 
     & \multicolumn{1}{c}{73.77} & \multicolumn{1}{c}{38.21} & \multicolumn{1}{c}{26.64} & \multicolumn{1}{c}{54.71} 
     & \multicolumn{1}{c}{86.28} & \multicolumn{1}{c}{42.64} & \multicolumn{1}{c}{28.27} & \multicolumn{1}{c}{59.07}
     & \multicolumn{1}{c}{70.63} & \multicolumn{1}{c}{35.69} & \multicolumn{1}{c}{25.51} & \multicolumn{1}{c}{52.28} \\

     \multirow{1}{*}{\textbf{Ours}} & \multicolumn{1}{c}{} 
     & \multicolumn{1}{c}{\textbf{89.72}} & \multicolumn{1}{c}{\textbf{44.56}} & \multicolumn{1}{c}{\textbf{28.96}} & \multicolumn{1}{c}{\textbf{62.13}}
     & \multicolumn{1}{c}{\textbf{83.14}} & \multicolumn{1}{c}{\textbf{42.17}} & \multicolumn{1}{c}{\textbf{27.92}} & \multicolumn{1}{c}{\textbf{59.44}} 
     & \multicolumn{1}{c}{\textbf{89.71}} & \multicolumn{1}{c}{\textbf{45.31}} & \multicolumn{1}{c}{\textbf{29.06}} & \multicolumn{1}{c}{\textbf{62.11}}
     & \multicolumn{1}{c}{\textbf{79.84}} & \multicolumn{1}{c}{\textbf{40.84}} & \multicolumn{1}{c}{\textbf{27.28}} & \multicolumn{1}{c}{\textbf{57.54}} \\

     \bottomrule
     \end{tabular}
     }
 \caption{Experimental results on the ScanRefer~\cite{chen2020scanrefer}. C, B-4, M, and R represent the captioning metrics CIDEr \cite{vedantam2015cider}, BLEU-4 \cite{papineni2002bleu}, METEOR \cite{banerjee2005meteor}, and ROUGE-L \cite{chin2004rouge}, respectively. A higher score for each indicates better performance.
 }
 \label{tab:scanrefer}
 \end{table*} 

\section{Experiments}
\label{sec:experiments}


\subsection{Datasets and Metrics}
\label{sec:datasets_and_metrics}

\paragraph{Datasets.}
In our studies, we focus on 3D dense captioning and employ two established datasets: ScanRefer \cite{chen2020scanrefer} and Nr3D \cite{achlioptas2020referit3d}.
These datasets are rich in human-generated descriptions, with ScanRefer providing $36,665$ descriptions for $7,875$ objects across $562$ scenes, and Nr3D offering $32,919$ descriptions for $4,664$ objects in $511$ scenes. 
For training, these descriptions and objects are derived from the ScanNet \cite{dai2017scannet} database, which comprises $1,201$ 3D scenes.
For evaluation, we use $9,508$ descriptions from ScanRefer and $8,584$ from Nr3D, corresponding to $2,068$ and $1,214$ objects across $141$ and $130$ scenes, respectively, from the $312$ 3D scenes in the ScanNet validation set.

\paragraph{Metrics.}
We evaluate the model using four types of performance metrics: CIDEr \cite{vedantam2015cider}, BLEU-4 \cite{papineni2002bleu}, METEOR \cite{banerjee2005meteor}, and ROUGE-L \cite{chin2004rouge}, denoted as \textbf{C}, \textbf{B-4}, \textbf{M}, and \textbf{R}, respectively.
Following the previous studies \cite{chen2021scan2cap, cai20223djcg, jiao2022more, wang2022spacap3d, chen2023vote2capdetr}, we first employ Non-Maximum Suppression (NMS) to eliminate redundant object predictions from the object proposals.
Each proposal is represented as a pair consisting of a predicted bounding box $\hat{b}_{i}$ and a generated caption $\hat{c}_{i}$.
To assess both the model's ability to locate objects and generate captions accurately, we employ the $m$@$k$, setting the IoU threshold $k$ at $0.25$ and $0.5$ for our experiments, following \cite{chen2021scan2cap}:
\begin{equation}
m @ k = \frac{1}{N} \sum_{i=1}^N m\left(\hat{c}_i, C_i\right) \cdot \mathbb{I}\left\{\operatorname{IoU}\left(\hat{b}_i, b_i\right) \geq k\right\},
\end{equation}
where $N$ is the number of all annotated instances in the evaluation set, and $m$ represents the captioning metrics C, B-4, M, and R.
\subsection{Implementation Details}
\label{sec:implementation_details}

Our training phase is structured into three stages, following the approach of \cite{chen2023vote2capdetr}. 
Initially, we pre-train our network excluding the caption head on the ScanNet~\cite{dai2017scannet} dataset for $1,080$ epochs.
The batch size is $8$.
The loss function is minimized using AdamW optimizer~\cite{adamw}, for which the initial learning rate is $5\times10^{-4}$ and decreases to $10^{-6}$ according to a cosine annealing schedule. 
We also apply a weight decay of $0.1$ and a gradient clipping of 0.1, as suggested by \cite{3detr}.
Afterward, starting from the pre-trained weights, we jointly train the entire model with the standard cross-entropy loss for an additional $720$ epochs on the ScanRefer~\cite{chen2020scanrefer} and Nr3D~\cite{achlioptas2020referit3d} datasets, fixing the detector's learning rate at $10^{-6}$ and reducing the caption head's from $10^{-4}$ to $10^{-6}$ to prevent over-fitting (about 20/17 hours for ScanRefer/Nr3D).
In the SCST~\cite{scst} phase, we adjust the caption head using a batch size of $2$ while keeping the detector fixed over a span of $180$ epochs and maintain a constant learning rate of $10^{-6}$ (about 22/18 hours for ScanRefer/Nr3D).
In the experimental setup that uses additional 2D data, as shown in \Cref{tab:scanrefer}, we employ the pre-trained ENet~\cite{enet} to extract $128$-dimensional multiview features from 2D ScanNet images, as in the \cite{chen2021scan2cap}.
The parameter size of our model is $21$M, and the average inference time on the evaluation set of the ScanRefer is $1.8$ms.
All experiments of our SIA are conducted with one Titan RTX GPU on PyTorch \cite{pytorch}.

\subsection{Comparison with Existing Methods}
\label{sec:comparison_with_existing_methods}

In this section, we benchmark our performance against eleven state-of-the-art methods: Scan2Cap~\cite{chen2021scan2cap}, D3Net~\cite{chen2022d3net}, SpaCap3D~\cite{wang2022spacap3d}, MORE~\cite{jiao2022more}, 3DJCG~\cite{cai20223djcg}, Contextual~\cite{zhong2022contextual}, REMAN~\cite{reman}, 3D-VLP~\cite{jin20233dvlp}, $\chi$-Tran2Cap~\cite{yuan2022xtrans2cap}, Vote2Cap-DETR~\cite{chen2023vote2capdetr}, and Unit3D~\cite{chen2023unit3d}. 
We apply IoU thresholds of $0.25$ and $0.5$ for ScanRefer~\cite{chen2020scanrefer} as shown in \Cref{tab:scanrefer} and an IoU threshold of $0.5$ for Nr3D~\cite{achlioptas2020referit3d} indicated in \Cref{tab:nr3d}.
For the baselines, we present the evaluation results reported in the original papers, and "-" in \Cref{tab:scanrefer} and \Cref{tab:nr3d} means that such results have not reported in the original paper or follow-up study.

\begin{table}
   \centering
   \resizebox{\linewidth}{!}{%
     \begin{tabular}{cc|cccc}
     \toprule
     \multirow{1}{*}{Model} & \multicolumn{1}{c}{Training} & \multicolumn{1}{c}{C@$0.5\uparrow$} & \multicolumn{1}{c}{B-4@$0.5\uparrow$} & \multicolumn{1}{c}{M@$0.5\uparrow$} & \multicolumn{1}{c}{R@$0.5\uparrow$} \\
     \midrule
     \midrule
     
     \multirow{1}{*}{Scan2Cap} & \multicolumn{1}{c}{} & \multicolumn{1}{c}{27.47} & \multicolumn{1}{c}{17.24} & \multicolumn{1}{c}{21.80} & \multicolumn{1}{c}{49.06} \\

     \multirow{1}{*}{D3Net} & \multicolumn{1}{c}{} & \multicolumn{1}{c}{33.85} & \multicolumn{1}{c}{20.70} & \multicolumn{1}{c}{23.13} & \multicolumn{1}{c}{53.38} \\

     \multirow{1}{*}{SpaCap3d} & \multicolumn{1}{c}{} & \multicolumn{1}{c}{33.71} & \multicolumn{1}{c}{19.92} & \multicolumn{1}{c}{22.61} & \multicolumn{1}{c}{50.50} \\

     \multirow{1}{*}{3DJCG} & \multicolumn{1}{c}{} & \multicolumn{1}{c}{38.06} & \multicolumn{1}{c}{22.82} & \multicolumn{1}{c}{23.77} & \multicolumn{1}{c}{52.99} \\

     \multirow{1}{*}{Contextual} & \multicolumn{1}{c}{MLE} & \multicolumn{1}{c}{35.26} & \multicolumn{1}{c}{20.42} & \multicolumn{1}{c}{22.77} & \multicolumn{1}{c}{50.78} \\

     \multirow{1}{*}{REMAN} & \multicolumn{1}{c}{} & \multicolumn{1}{c}{34.81} & \multicolumn{1}{c}{20.37} & \multicolumn{1}{c}{23.01} & \multicolumn{1}{c}{50.99} \\

     \multirow{1}{*}{Vote2Cap-DETR} & \multicolumn{1}{c}{} & \multicolumn{1}{c}{43.84} & \multicolumn{1}{c}{26.68} & \multicolumn{1}{c}{25.41} & \multicolumn{1}{c}{54.43} \\

     \multirow{1}{*}{\textbf{Ours}} & \multicolumn{1}{c}{} & \multicolumn{1}{c}{\textbf{56.39}} & \multicolumn{1}{c}{\textbf{30.87}} & \multicolumn{1}{c}{\textbf{27.54}} & \multicolumn{1}{c}{\textbf{60.36}} \\
     
     \midrule

     \multirow{1}{*}{D3Net} & \multicolumn{1}{c}{} & \multicolumn{1}{c}{38.42} & \multicolumn{1}{c}{22.22} & \multicolumn{1}{c}{24.74} & \multicolumn{1}{c}{54.37} \\

     \multirow{1}{*}{$\chi$-Tran2Cap} & \multicolumn{1}{c}{} & \multicolumn{1}{c}{33.62} & \multicolumn{1}{c}{19.29} & \multicolumn{1}{c}{22.27} & \multicolumn{1}{c}{50.00} \\

     \multirow{1}{*}{Contextual} & \multicolumn{1}{c}{SCST} & \multicolumn{1}{c}{37.37} & \multicolumn{1}{c}{20.96} & \multicolumn{1}{c}{22.89} & \multicolumn{1}{c}{51.11} \\
          
     \multirow{1}{*}{Vote2Cap-DETR} & \multicolumn{1}{c}{} & \multicolumn{1}{c}{45.53} & \multicolumn{1}{c}{26.88} & \multicolumn{1}{c}{25.43} & \multicolumn{1}{c}{54.76} \\

     \multirow{1}{*}{\textbf{Ours}} & \multicolumn{1}{c}{} & \multicolumn{1}{c}{\textbf{59.48}} & \multicolumn{1}{c}{\textbf{32.60}} & \multicolumn{1}{c}{\textbf{27.99}} & \multicolumn{1}{c}{\textbf{61.08}} \\
     
     \bottomrule
     \end{tabular}
     }
 \caption{Experimental results on the Nr3D~\cite{achlioptas2020referit3d} with IoU threshold at $0.5$. 
 }
 \label{tab:nr3d}
 \end{table} 
\paragraph{ScanRefer.}
The descriptions in the ScanRefer include depictions of the target object's attributes and information about the spatial relationships between this target object and other surrounding objects.
\Cref{tab:scanrefer} summarizes the results on the ScanRefer dataset.
Our method significantly surpasses current methods in 3D dense captioning across all input data settings and IoU threshold configurations.
We attribute this enhancement to our contextualized late aggregation mechanism.
Unlike previous object-centric methods where surrounding information is bound to the center of object proposals, SIA directly targets the region-of-interest for contextual captions with the context query while simultaneously addressing object localization and its attribute captions with the instance query.

\paragraph{Nr3D.}
The Nr3D dataset is designed to evaluate the model's performance in interpreting free-form natural language descriptions of objects as spoken by humans. 
Our SIA quantitatively demonstrates its ability to generate various descriptions for an object by showing state-of-the-art performance across all evaluation metrics, as shown in \Cref{tab:nr3d}.

\begin{table}[t]
   \centering
   \resizebox{\linewidth}{!}{%
     \begin{tabular}{ccccccc}
     \toprule
     \multirow{1}{*}{} & \multicolumn{6}{c}{IoU=$0.50$} \\
     \cmidrule(lr){2-7}
     \multirow{1}{*}{Model} & \multicolumn{1}{c}{C$\uparrow$} & \multicolumn{1}{c}{B-4$\uparrow$} & \multicolumn{1}{c}{M$\uparrow$} & \multicolumn{1}{c}{R$\uparrow$} & \multicolumn{1}{c}{mAP$\uparrow$} & \multicolumn{1}{c}{AR$\uparrow$} \\
     \midrule
     \midrule

     \multirow{1}{*}{Vote2Cap-DETR}
     & \multicolumn{1}{c}{73.77} & \multicolumn{1}{c}{38.21} & \multicolumn{1}{c}{26.64} & \multicolumn{1}{c}{54.71} & \multicolumn{1}{c}{45.56} & \multicolumn{1}{c}{67.77} \\


     \multirow{1}{*}{SIA using only $V^o$}
     & \multicolumn{1}{c}{73.90} & \multicolumn{1}{c}{40.67} & \multicolumn{1}{c}{26.76} & \multicolumn{1}{c}{55.31} & \multicolumn{1}{c}{48.09} & \multicolumn{1}{c}{68.43} \\


     \multirow{1}{*}{SIA w/o $V^g$}
     & \multicolumn{1}{c}{81.45} & \multicolumn{1}{c}{41.19} & \multicolumn{1}{c}{26.33} & \multicolumn{1}{c}{56.71} & \multicolumn{1}{c}{48.74} & \multicolumn{1}{c}{68.13} \\

     \multirow{1}{*}{\textbf{SIA (Ours)}}
     & \multicolumn{1}{c}{\textbf{83.14}} & \multicolumn{1}{c}{\textbf{42.17}} & \multicolumn{1}{c}{\textbf{27.92}} & \multicolumn{1}{c}{\textbf{59.44}} & \multicolumn{1}{c}{\textbf{49.69}} & \multicolumn{1}{c}{\textbf{69.08}} \\

     \bottomrule
     \end{tabular}
     }
 \caption{Ablation study on the ScanRefer~\cite{chen2020scanrefer}. 
 The core components of SIA: i) decomposing the query set into the instance query $V^o$ and the context query $V^c$, ii) generating the global feature $V^g$, and iii) aggregating the TGI feature $V^a$.
 }
 \label{tab:ablation_key_components}
 \end{table} 
\begin{table}
   \centering
   \resizebox{\linewidth}{!}{%
     \begin{tabular}{c|cccc}
     \toprule
     \multirow{1}{*}{Method} & \multicolumn{1}{c}{C@$0.5\uparrow$} & \multicolumn{1}{c}{B-4@$0.5\uparrow$} & \multicolumn{1}{c}{M@$0.5\uparrow$} & \multicolumn{1}{c}{R@$0.5\uparrow$} \\
     \midrule
     \midrule

     \multirow{1}{*}{Contexts $V^c$} & \multicolumn{1}{c}{72.82} & \multicolumn{1}{c}{38.46} & \multicolumn{1}{c}{26.76} & \multicolumn{1}{c}{56.71} \\
          
     \multirow{1}{*}{Single Instance $V^o_i$ \& Contexts $V^c$} & \multicolumn{1}{c}{72.89} & \multicolumn{1}{c}{38.21} & \multicolumn{1}{c}{27.04} & \multicolumn{1}{c}{57.33} \\

     \multirow{1}{*}{\textbf{Instances $V^o$ \& Contexts $V^c$ (Ours)}} & \multicolumn{1}{c}{\textbf{73.22}} & \multicolumn{1}{c}{\textbf{40.91}} & \multicolumn{1}{c}{\textbf{28.19}} & \multicolumn{1}{c}{\textbf{60.46}} \\
     
     \bottomrule
     \end{tabular}
     }
 \caption{Ablation for how the instance feature $V^o$ and the context features $V^c$ participate in the Global Aggregator $G(\cdot)$ on the ScanRefer~\cite{chen2020scanrefer}. $\cdot_i$ denotes a single $i$-th decoded query feature.}
 \label{tab:ablation_global_feature}
 \end{table} 
\begin{table}[t]
   \centering
   \resizebox{\linewidth}{!}{%
     \begin{tabular}{c|cccc}
     \toprule
     \multirow{1}{*}{} & \multicolumn{1}{c}{C@$0.5\uparrow$} & \multicolumn{1}{c}{B-4@$0.5\uparrow$} & \multicolumn{1}{c}{M@$0.5\uparrow$} & \multicolumn{1}{c}{R@$0.5\uparrow$} \\
     \midrule
     \midrule

     \multirow{1}{*}{SIA with GeM Pooling} & \multicolumn{1}{c}{66.97} & \multicolumn{1}{c}{36.97} & \multicolumn{1}{c}{26.76} & \multicolumn{1}{c}{56.71} \\
          
    \multirow{1}{*}{\textbf{SIA with NetVLAD (Ours)}} & \multicolumn{1}{c}{\textbf{73.22}} & \multicolumn{1}{c}{\textbf{40.91}} & \multicolumn{1}{c}{\textbf{28.19}} & \multicolumn{1}{c}{\textbf{60.46}} \\
    
     \bottomrule
     \end{tabular}
     }
 \caption{Experimental results comparing Global Aggregators on the ScanRefer~\cite{chen2020scanrefer}.}
 \label{tab:global_pooling}
 \end{table} 
\subsection{Ablation Study and Discussion}
\label{sec:ablation_study}


The core components of SIA consist of three factors: i) decomposing the query set into instance query $V^o$ and context query $V^c$, ii) generating the global descriptor $V^g$, and iii) composing the fully informed contextualized feature $V^a$ using our TGI-Aggregator.
In our ablation study, we validate that every component of our proposed SIA positively contributes to the final performance.

\paragraph{Instance Query Generator.}
We define SIA using only the Instance Query Generator (i.e., SIA using only $V^o$ in \Cref{tab:ablation_key_components}) as our baseline and compare it with Vote2Cap-DETR~\cite{chen2023vote2capdetr}, an object-centric transformer encoder-decoder architecture.
The major difference between our baseline and Vote2Cap-DETR is how we generate the query set for instances.
Vote2Cap-DETR uses farthest point sampling (FPS) to generate queries before the query coordinates are adjusted through voting.
Therefore, if the coordinates are mistakenly focused on a specific object after voting, features will be extracted from the same object.
On the other hand, our baseline extracts the features from the candidate coordinates after the voting. 
This enhancement boosts localization performance in terms of mean Average Precision (mAP) and Average Recall (AR), naturally leading to improvements in dense captioning performance.

\paragraph{Context Query Generator.}
SIA decomposes the role of queries into instance query $V^o$ that focuses on the object itself and context query $V^c$ that designates the contextual region.
SIA w/o $V^g$ in \Cref{tab:ablation_key_components} shows the result of generating a caption using only context query $V^c$ and instance query $V^o$ in the TGI-aggregator, excluding the global descriptor.
By showing high performance improvement compared to the results of SIA using only $V^o$, we demonstrate the effectiveness of query set separation.

\paragraph{TGI-Aggregator.}
As shown in \Cref{tab:ablation_key_components}, utilizing a global descriptor $V^g$ in the TGI-Aggregator results in performance improvements across all aspects.
\Cref{tab:ablation_global_feature} shows our ablation study on how we aggregate the decoded context query $V^c$ and instance query $V^o$ to construct the global feature $V^g$ that is afterward fed to our TGI-Aggregator (recall \Cref{fig:tgi}).
The scenarios include i) aggregating all context features $V^c$, ii) gathering one instance feature $V^o_i$ with all context features $V^c$, and iii) aggregating all instance features $V^o$ and context features $V^c$ to extract a global feature.
Aggregating all instance and context features to create a global feature results in the best performance.
This implies that reflecting all instances and contexts is better when representing the entire scene.

We also compare two of the most widely used aggregation frameworks for whole-scene representation: GeM pooling~\cite{gem} and NetVLAD~\cite{netvlad}, as shown in \Cref{tab:global_pooling}.
We empirically adopt NetVLAD for our Global Aggregator. 
In our TGI-Aggregator, we concatenate the global descriptor $V^g$ to each decoded instance query $V^o_i$ and $K$ context features that are locationally closest to $V^o_i$ within $V^c$.


\begin{table}
   \centering
   \resizebox{\linewidth}{!}{%
     \begin{tabular}{c|cccc}
     \toprule
     \multirow{1}{*}{} & \multicolumn{1}{c}{C@$0.5\uparrow$} & \multicolumn{1}{c}{B-4@$0.5\uparrow$} & \multicolumn{1}{c}{M@$0.5\uparrow$} & \multicolumn{1}{c}{R@$0.5\uparrow$} \\
     \midrule
     \midrule

     \multirow{1}{*}{SIA with K=8} & \multicolumn{1}{c}{69.86} & \multicolumn{1}{c}{37.89} & \multicolumn{1}{c}{27.04} & \multicolumn{1}{c}{57.33} \\

     \multirow{1}{*}{\textbf{SIA with K=16 (Ours)}} & \multicolumn{1}{c}{73.22} & \multicolumn{1}{c}{\textbf{40.91}} & \multicolumn{1}{c}{\textbf{28.19}} & \multicolumn{1}{c}{\textbf{60.46}} \\
     
     \multirow{1}{*}{SIA with K=32} & \multicolumn{1}{c}{\textbf{73.38}} & \multicolumn{1}{c}{37.92} & \multicolumn{1}{c}{27.92} & \multicolumn{1}{c}{60.16} \\
     
     \bottomrule
     \end{tabular}
     }
 \caption{Performance variation according to the size of K on the ScanRefer~\cite{chen2020scanrefer}.}
 \label{tab:analysis_K}
 \end{table} 
\paragraph{Analysis of hyper-parameter K.}
\label{sec:analysis_k}
In the TGI-Aggregator, we concatenate the global descriptor $V^g$ to each decoded instance query $V^o_i$ and $K$ nearest features within $V^c$ in terms of spatial proximity to $V^o_i$.
Based on the experimental results of \Cref{tab:analysis_K}, we set $K=16$. 
A performance decrease is observed with $K=8$, likely due to insufficient contextual information, while $K=32$ shows little performance change despite significantly increasing memory and execution time costs.

\begin{table}[t]
\centering

\begin{subfigure}{\linewidth}
\centering
\tiny
\begin{tabular}{c|p{.7\linewidth}}
\toprule
Object & scene0144\_00 | 20 | nightstand \\
\midrule

Only use $V^o$ & "this is a brown nightstand ." \\
\midrule
Only use $V^a$ & "it is to the left of the bed ." \\
\midrule
SIA & "this is a brown nightstand . it is to the left of the bed ." \\
\midrule
GT & "there is a nightstand on the wall . it is to the left of a bed ." \\
\bottomrule
\end{tabular}
\label{fig:table_a}
\end{subfigure}
\hfill

\begin{subfigure}{\linewidth}
\centering
\tiny
\begin{tabular}{c|p{.7\linewidth}}
\toprule
Object & scene0019\_00 | 9 | vending\_machine \\
\midrule

Only use $V^o$ & "the vending machine is a white rectangle ." \\
\midrule
Only use $V^a$ & "the vending machine is in the corner of the room ." \\
\midrule
SIA & "the vending machine is in the corner of the room . the vending machine is a white rectangle ." \\
\midrule
GT & "this is a vending machine . it is in the corner of the room , by a lamp ." \\
\bottomrule
\end{tabular}
\label{fig:table_a}
\end{subfigure}
\hfill

\caption{Qualitative results on the ScanRefer~\cite{chen2020scanrefer} generated from i) using only instance features $V^o$, ii) using the fully informed contextualized feature $V^a$ from the TGI-Aggregator, and iii) concatenating both captions for the final caption of SIA.
}
\label{tab:samples}
\end{table}

\subsection{Qualitative Analysis}
\label{sec:qualitative_analysis}
\begin{figure*}[t]
\begin{center}
\centerline{\includegraphics[width=\textwidth]{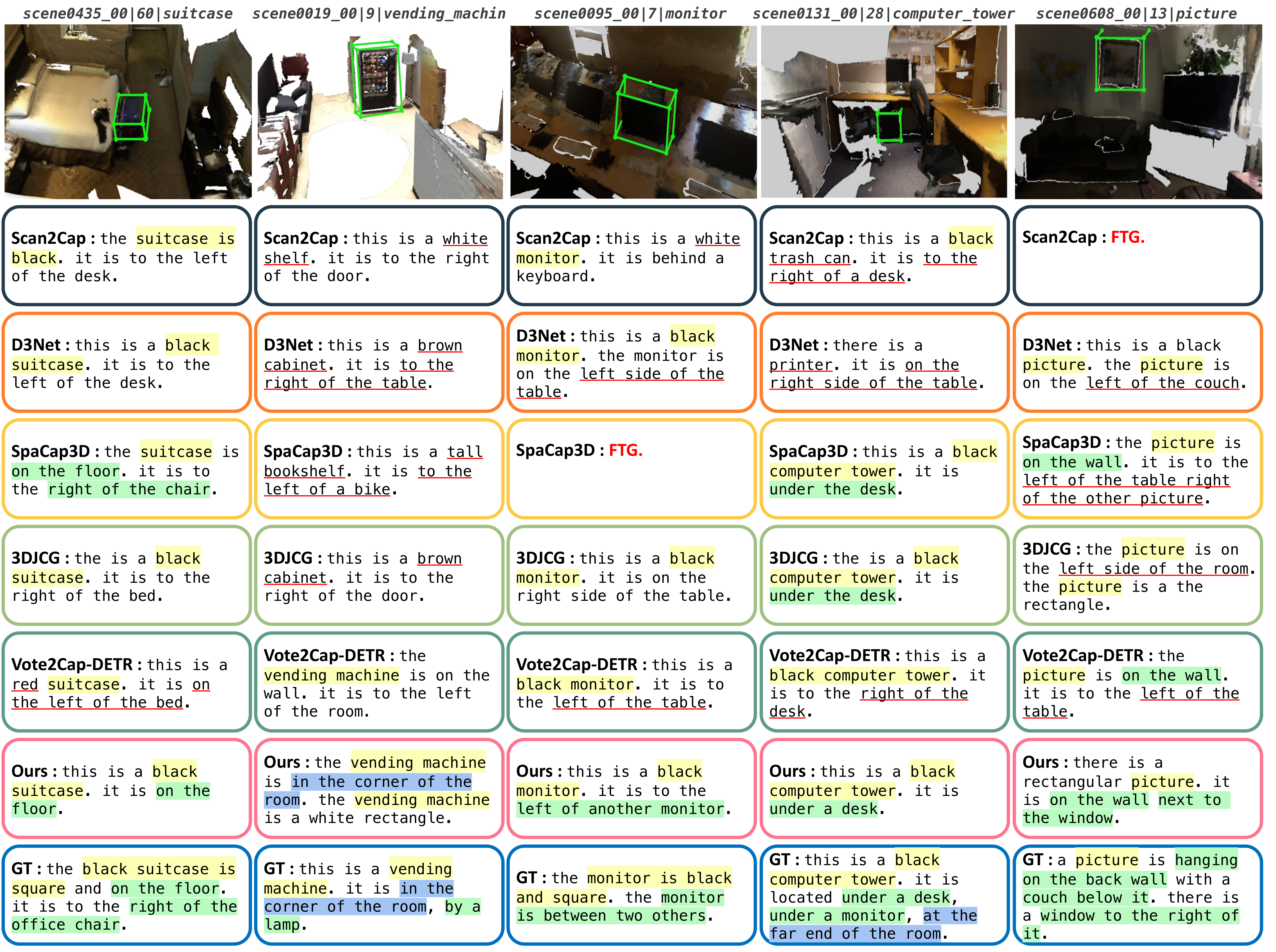}}
\caption{
Qualitative results on the ScanRefer \cite{chen2020scanrefer}. 
The yellow-highlighted sections show information specific to the object itself, the green-highlighted sections describes the relationships between objects, and the blue-highlighted sections depict the spatial position of the object in the 3D scene. 
Captions underlined in red indicate incorrect descriptions. 
\textbf{FTG.} represent failures in caption generation due to low IoU.
}
\label{fig:qualitative_results}
\end{center}
\end{figure*}

We qualitatively present the captions generated from i) using only instance features $V^o$, ii) using only the fully informed contextualized feature $V^a$ from the TGI-Aggregator, as shown in \Cref{tab:samples}.
It can be seen that while the captions generated from $V^o$ include descriptions for attributes, the captions generated from $V^a$ include contextual information such as relations with other objects and the global scene.

We also provide a qualitative comparison with the state-of-the-art models: 
Scan2Cap~\cite{chen2021scan2cap}, SpaCap3D~\cite{wang2022spacap3d}, D3Net~\cite{chen2022d3net}, 3DJCG~\cite{cai20223djcg}, and Vote2Cap-DETR~\cite{chen2023vote2capdetr}. 
The ground-truth includes descriptions of the intrinsic properties of the object (e.g., the monitor is black and square.), explanations using the relationships between objects (e.g., a picture is hanging on the back wall with a couch below it.), and descriptions of the object in the context of the entire space (e.g., at the far end of the room.).
Existing models, focusing on objects, generate captions limited to the object and its immediate relations in a fixed format (e.g., this is a black suitcase.).
Our model can handle not only object queries but also context queries, allowing it to generate sentences in various formats (e.g., the vending machine is a white rectangle.) and create descriptions of the entire space (e.g., the vending machine is in the corner of the room.).
These results emphasize the importance of an integrated understanding of the object, its surroundings, and the overall space in captioning.

\section{Conclusion}

In the 3D dense captioning task, the description of an object within a 3D scene encompasses not only the intrinsic characteristics of the object but also the relationship with surrounding objects and the spatial relationship of the object with respect to the overall space.
In this work, we propose a novel approach that independently generates captions with different region of interests and aggregates them afterwards to enhance local-global sensitivity of descriptions.
Through extensive experiments on benchmark datasets, our method significantly improves 3D dense captioning over previous approaches, demonstrating the importance of an integrated understanding of objects, surroundings, and overall space for caption generation.

\section*{Limitations}
SIA still faces the limitations of prior set prediction architectures, where the number of instance and context queries must be heuristically determined.
Future work could explore methods that allow for the dynamic adjustment of instances and context query numbers based on the complexity of the scene.


\section*{Ethical Considerations}
3D dense captioning is the task of locating objects in a 3D scene and generating captions for each object.
Our proposed method, specifically designed for this task, is ethically sound, employing only publicly available datasets throughout our research. 
These benchmark datasets feature 3D indoor environments exclusively populated with objects.

\section*{Acknowledgement}
We sincerely thank Junhyug Noh, Seokhee Hong, Jaewoo Ahn, and Dayoon Ko for their constructive comments.
This work was supported by LG AI Research and Institute of Information \& Communications Technology Planning \& Evaluation (IITP) grant (No.~RS-2019-II191082, No.~RS-2022-II220156) funded by the Korea government (MSIT).

\bibliography{anthology,custom}


\end{document}